\documentclass[10pt, a4paper]{article}
\usepackage{comment}
\usepackage[]{lrec2026} 
\usepackage{xcolor}

\title{Dutch Metaphor Extraction from Cancer Patients' Interviews and Forum Data using LLMs and Human in the Loop}

\name{Lifeng Han$^{1,2}$, David Lindevelt$^{2}$, Sander Puts$^{3}$, Erik van Mulligen$^{4}$,  Suzan Verberne$^{2}$\\ on behalf of 4D PICTURE} 

\address{ $^1$Biomedical Data Sciences, Leiden University Medical Center, NL \\ $^2$Leiden Institute of Advanced Computer Science (LIACS), Leiden University, NL\\
$^3$Department of Radiation Oncology (MAASTRO), GROW Research Institute for Oncology and Reproduction,\\ Maastricht University Medical Centre+, Maastricht, NL\\
$^4$Department of Medical Informatics, Erasmus University Medical Center Rotterdam, NL\\
         \{l.han, s.verberne\}@liacs.leidenuniv.nl\\}

\abstract{
Metaphors play an important role in communication between cancer patients and clinicians. In this work, we present the first study on automated extraction of Dutch metaphors from cancer patient interviews and online forum data using large language models (LLMs) and human validation.
We develop a human-in-the-loop extraction framework combining structured prompting, automatic verification, and expert linguistic evaluation. We evaluate multiple open-source LLMs and prompting strategies, including chain-of-thought and knowledge-guided prompting.
Our approach resulted in the HealthQuote.NL dataset, containing 130 validated metaphors. We analyse common model errors, including hallucination and confusion with idioms, and show that refined prompting achieved 63.2\% precision.
This work provides both a novel dataset and an extraction framework to support patient-centred healthcare communication and future metaphor research.
We share our prompts and related resources at \url{https://github.com/4dpicture/HealthQuote.NL}.
 \\ \newline  }

\begin{document}

\maketitleabstract


\section{Introduction}
According to the Contemporary theory of metaphor, a metaphor is ``a cross-domain mapping in the conceptual system'' \cite{lakoff1993contemporary}. Metaphor can be expressed in different linguistic units, such as words, phrases, or sentence levels. People use metaphors in daily life with or without realising them.

In the healthcare domain, metaphors can play a helpful role in the communication between caregivers and cancer patients~\cite{semino2017metaphor,harrington2012use,liu2024cancer}: patients can be helped by comparing their cancer treatment process to e.g. journeys or battles. In the UK, the Metaphor Menu has been a successful project, publishing a pamphlet of potentially helpful metaphors for patients living with cancer.\footnote{\url{https://wp.lancs.ac.uk/melc/the-metaphor-menu/}} 

In this paper, we take the first steps in collecting a set of Dutch-language metaphors related to living with cancer. We extract these metaphors from patient utterances, both spoken (transcribed) and written. To this end, we analyse two distinct corpora of Dutch patient narratives: interviews and blog posts (kanker.nl). We experiment with a suite of small, local LLMs and a range of prompting techniques to extract metaphors from texts. After analysing the output with simple instruction prompts, we identify 7 challenges for this task, including hallucination of metaphors, figurative language or idioms mistaken as metaphors, and abstraction instead of extraction of the originally mentioned metaphors. With more advanced prompting techniques, including chain-of-thought and iterative self-prompting, we were able to address these challenges to some extent and extract useful metaphors from the two datasets.


In summary, our contributions are:
\begin{itemize}
\item We present the first study investigating large language models for metaphor extraction in Dutch cancer patient narratives.
\item We develop a human-in-the-loop extraction framework combining structured prompting, automatic verification, and expert linguistic validation.
\item We construct HealthQuote.NL, a curated dataset of 130 validated Dutch metaphors derived from interviews and forum data.
\item We provide an empirical analysis of prompting strategies, identifying common failure modes such as hallucination, abstraction, and idiom confusion.
\item We demonstrate the potential of LLM-assisted metaphor extraction to support patient-centred healthcare communication.\footnote{Regarding data sharing of HealthQuote.NL, to protect patients' privacy, we generate paraphrased and synthetic examples using the metaphors we extracted from online forum text.}

\end{itemize}

\section{Related Work}
In this section, we introduce works from three perspectives: metaphor identification in Dutch, metaphors in cancer, and metaphors in Dutch healthcare.

\subsection{Metaphor Identification in Dutch}
Regarding metaphor identification and extraction methods in Dutch, to the best of our knowledge, little work has been done.
One PhD thesis by \newcite{pasma2011metaphor_PhD} focused on metaphor and register variation in Dutch news discourse.
They extended the metaphor identification procedure (MIP) from \newcite{group2007mip} by adapting it to some new situation when ``cross-domain mapping'' appears for certain words, including ``direct metaphor'', ``implicit metaphor'', and ``metaphor flag''. 
This adaptation has been applied to the Dutch discourse corpus on news texts.
The MIP principle is a symbolic text examination on a word-by-word basis.
Later on, there have been a few works published as book chapters (non-open-access books) by the same research group, following the PhD thesis project, including \cite{pasma2012metaphor,pasma2019chapter}.

Another loosely related work is from \newcite{de2014conceptual}, which studies the conceptual metaphors of Dutch and German for efficient teaching and, alternatively, aims at verifying them.

\textit{Our work will be the first exploring and investigating LLMs for Dutch metaphor identification and extraction.}



\subsection{Metaphors in Cancer}
Various researchers have conducted metaphor studies within the cancer domain. Review articles in this topic include \newcite{harrington2012use,liu2024cancer}. 
\newcite{semino2017online} carried out comparative studies on metaphors used by cancer patients vs health professionals at online forums in the UK between 2007 and 2013. It included 56 patients and 307 health professionals. 
The study showed that ``violence'' and ``journey'' related metaphors are not necessarily negative or positive by default. It depends on how they are used. The ``violence'' metaphors were further studied and discussed in their following work on communicating nuanced results in Language Consultancy \cite{demjen2020communicating}.
Similar work was conducted by \newcite{declercq2023machines} on metaphors use of machines, journeys, and prisons for chronic pain patients in consultations to describe pain, illness, and medicines. 

There are courses designed, e.g. ``Cancer as Metaphor’’ by \newcite{penson2004cancer} for explaining the strengths and weaknesses of using metaphor in patient-caregiver relationships. The instructions covered different roles, including facilitator, oncologist, psychiatrist, and psychologist.

Focusing on the staging, researchers studied ``imprisonment'', ``burden'', ``battle’’, and ``journey’’ metaphors on patients living with \textit{advanced cancer} using blog posts \cite{hommerberg2020battle}, and a focus group design with eighteen people
on cancer \textit{survivor} \cite{appleton2014searching}.

From a gender perspective, \newcite{gibbs2002embodied} studied the relation between language use and thoughts from 6 females who live with cancer. 
From cancer types, \newcite{pfeifer2024narratives} conducted a study on how metaphors can help the communication of information exchange from breast cancer patients about depressive symptoms. 

There are also focused studies on language or region-specific, e.g. Spanish (online forum) \cite{magana2018spanish}, the UK (corpus-based) \cite{semino2017metaphor},
American and Nigerian (self-help books) on ``military, journey, personification, and sports'' \cite{nwankwo2024metaphor}.


Most work has focused on one source of data, e.g. either online data, interviews, or books. However, we aim to \textit{explore the difference between verbal and online-written data}. 
Similar to \cite{semino2017metaphor}, in the interview data we have, there are 3 roles (patient, significant other, and interviewer), but in Dutch (the Netherlands) instead of English (UK).



\subsection{Metaphors in Dutch Healthcare}
To our best knowledge, there is no prior work on Dutch-language metaphors in the cancer domain. However, there is some work on metaphor-related research in other Dutch healthcare domains that we list here.

To better understand the stigma toward \textit{dementia} patients, 
\newcite{creten2022stigma} conducted sentiment analysis of Dutch language using tweets from 2019 to 2020 into seven dimensions, including joke and metaphor, in addition to others such as information, organization, personal experience, politics, and ridicule. The authors identified 10\% of the tweets were metaphorical during manual coding.

To study the needs of \textit{hypertensive} patients expressing themselves using metaphors, analogies, and symbols, such as the causes and consequences of the illness, \newcite{schuster2011metaphorical} used 55 patients of three ethnic groups in the Netherlands, including Creole, Dutch, and Hindustani descent, for in-depth interviews.
The finding shows that patients use machine and enemy-related metaphors to explain the origins and consequences of hypertension, with its characteristics of silence, invisibility, and uncontrollability.

In another study by \newcite{hillen2025clinicians}, the authors investigated the usage of metaphoric language by clinicians for family members of \textit{ICU} patients. 
They sampled 36 neonatal, pediatric, and adult patients, together with 104 of their family members. 
Using the transcribed and anonymised audio recording of conversations, the findings show that metaphors were used frequently and predominantly by clinicians to share clinical information and the awareness of using metaphors by clinicians can improve effective information exchange. 
However, the authors also discussed that both harms and benefits arise from clinicians using metaphors, e.g. shared decision making (SDM) might be hindered if clinicians actively and consciously use metaphor to steer the families into a direction they consider as the best for patients.

In contrast to this, in our work, we aim to \textit{identify and extract metaphors used by \textbf{patients} and their \textbf{family} members} (significant others), for better support of patient care.





\begin{figure*}[!t]
\begin{center}
\includegraphics[width=\textwidth]{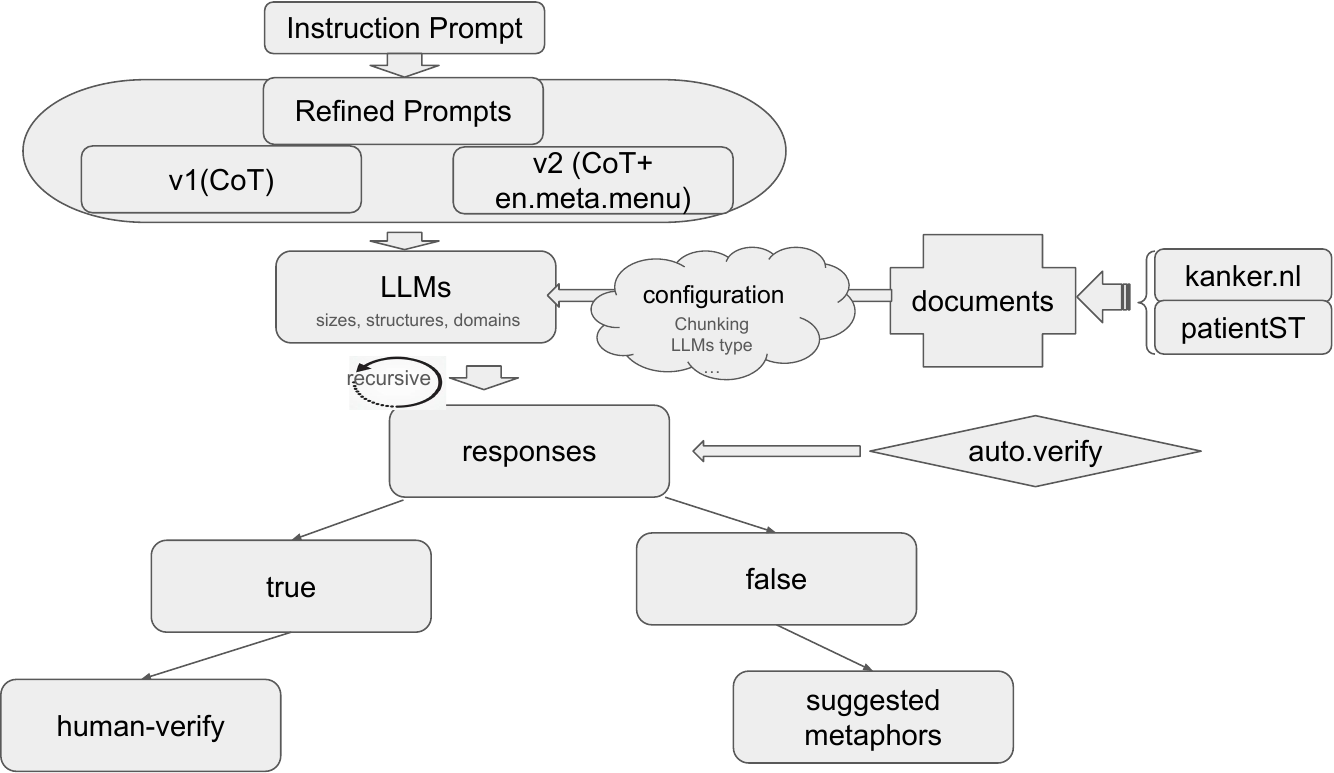}
\caption{\textsc{HealthQuote.NL}: Extraction Framework using LLMs and Human in the Loop. 
}
\label{fig.framework}
\end{center}
\end{figure*}

\section{Methods Design and Development}
\label{sec:method-design}
To identify metaphors used in texts in patient records, we explore different kinds of techniques on prompting LLMs, including Persona \cite{lutz2025promptmakespersonasystematic}, chain of thought (CoT) \cite{saparov2023language-CoT}, iterative self-prompting (ISP, \textit{aka} LLM-assisted prompting) \cite{romero-etal-2025-manchester,ren2025malei}, and knowledge-inserted prompting \cite{dong-etal-2024-survey}. 
To analyse if the structure, size of LLMs, and their pre-trained domain knowledge matter, we use LLMs from different sizes and structures, as well as domain-specific ones, including models pre-trained on medical knowledge.

As shown in Figure \ref{fig.framework}, the initial prompt we used is the ``Instruction Prompt (\textbf{I.inP})'' using Persona, role of ``system'' and ``user'', without CoTs. Messages we used for prompting are:

\begin{itemize}
    \item "role": "system", "content": "You are an assistant that processes and analyses documents."
    \item "role": "user", "content": "Here is the document: <\textit{document-content}> Please tell me metaphors that people used in the conversation in this document and list them in bullet points. Can you offer the English translation version too? Can you also offer the full sentences in which the metaphors were used?"
\end{itemize}

Then, in the refined prompts (\textbf{RPs}), we added CoT and ISP by asking LLMs to generate prompts for such a task. After a few iterations of refinement, the LLM-assisted long and structured prompt message for this task includes the following key components\footnote{We share the detailed prompts at a public GitHub page \url{https://github.com/aaronlifenghan/HealthQuote.NL}}:
\begin{itemize}
    \item Persona: an expert linguist
    \item Strict Extraction Protocol
    \begin{itemize}
        \item Document Scanning
        \item Metaphor Identification, CoTs with examples
        \item Verification Requirements, different levels
        \item Data Extraction Format
        \item Structured Output (Population)
    \end{itemize}
    \item Quality Control Rules
    \item Final Instructions
\end{itemize}

Among the Refined Prompts (RP v1 and v2), the difference is that for v1, we only gave three simple and short metaphorical sentences for few-shot prompting:
\begin{itemize}
    \item "This treatment is a journey" (medical process → travel)
    \item "Fighting cancer" (disease → war)
    \item "The tumor is growing like a weed" (cancer → plant)
\end{itemize}
For v2, we inserted the entire English Metaphor Menu list (17 from the public website \citetlanguageresource{engMetaMenu}) with our categorisation.
The different settings for RP-v1 and RP-v2 are to investigate 1) if the English Metaphor Menu can guide the LLMs to extract better metaphors; 2) if the English Metaphor Menu will introduce bias to LLMs to extract only metaphors like the English metaphors.

For automatic verification (auto.verify), we implemented an external checklist tool for LLMs: if they are able to find the exact text in the document that expresses the metaphor, identify the exact section, identify the speaker's role (when there are multiple roles), and identify the metaphor (not literal medical terms).

For extracted metaphors, we ask LLMs to group them into 1) types: word, phrase, sentence, and extended, 2) source domain: violence, journey, nature (garden), games, music, fairground, unwanted guests, religion, control, building, machine, and other, and 3) function in context: explanation, coping, empowerment, relationship, prognosis, treatment, emotion, and humour.




For security and privacy reasons on processing patients' data, even though they are already anonymised, we prioritise to use local LLMs through the ollama platform\footnote{\url{https://ollama.com/library}} over commercial models. 
The open-source models we used include both medical and general domains. 



\section{Experiments}

\subsection{Data}
\begin{figure}[!t]
\begin{center}
\includegraphics[width=0.49\textwidth]{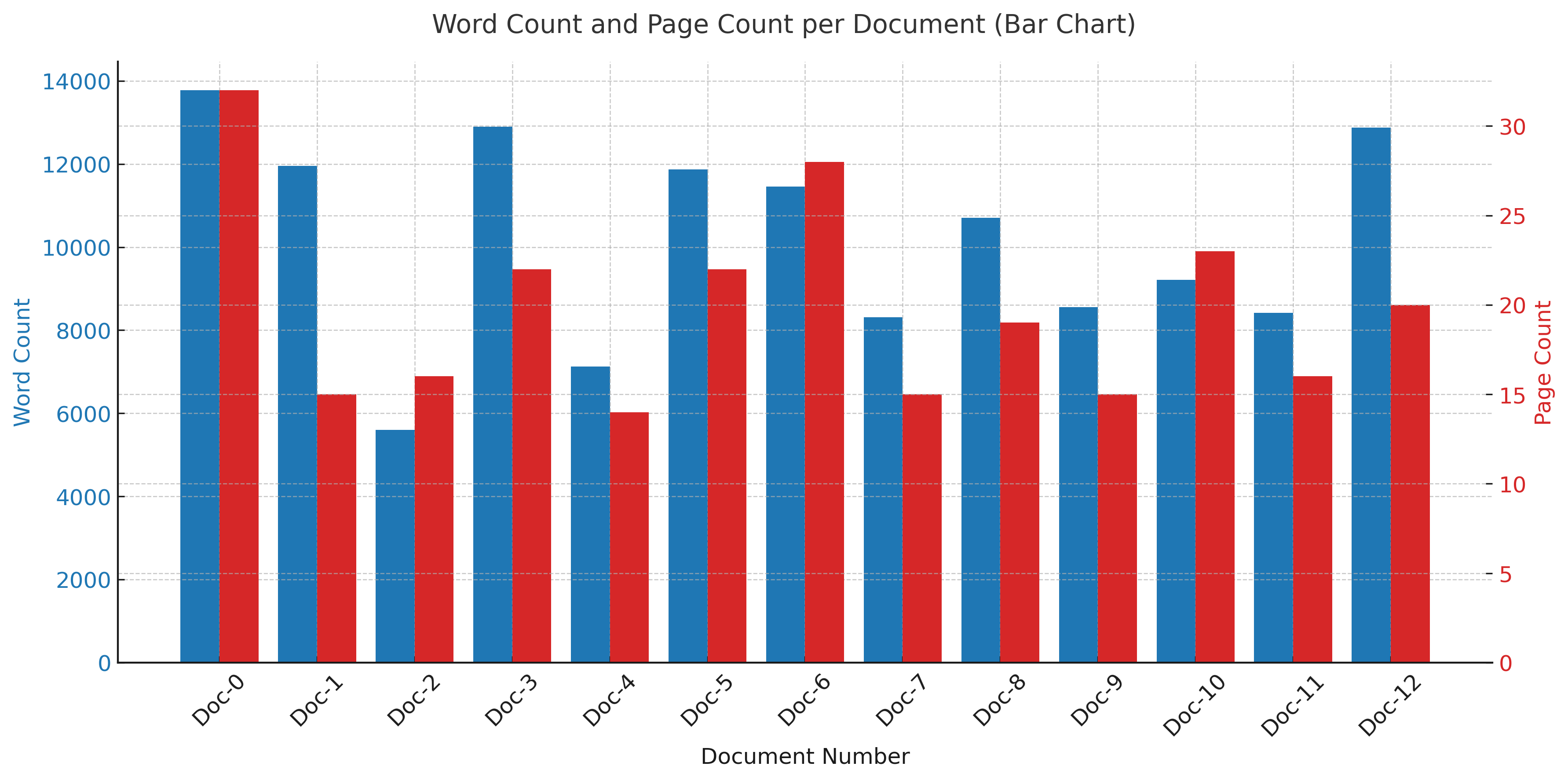}
\caption{Interview Data Statistics. 
}
\label{fig.interviewData}
\end{center}
\end{figure}

For \textbf{Patient Story Telling (patientST) Interview data}, we used the transcriptions of 13 documents from 13 interviews in an oncology setting obtained from \cite{griffioen2021bigger}. There are three roles for each interview: cancer patient, a family member or friend of the patient (significant other), and interviewer (researcher). 
They are specified as P (Patiënt), N (Naaste), and O (Onderzoeker) in the free text.
The largest document has  13,777 words and 32 pages, while the smallest document has 5,596 words and 16 pages.
See Figure \ref{fig.interviewData} for statistics per document.

For \textbf{cancer patient forum data}, we have a collection in Dutch of 15,653 blogs, 17,290 comments, 2,246 group discussions, 5,777 ``ask a professional'' items in which experts answer questions from patients, and 10,134 reactions on ``ask a professional'' from three cancer types: breast cancer (3,524), prostate cancer (2,613), and melanoma (614). The patient content includes treatment phases, treatments, and long-term consequences.

We have approval to use both the two sets of data for research. We can not disclose the original forum text. Considering this, instead of displaying the original text, we will give paraphrases or synthetic example texts generated from the extracted metaphors.



\subsection{Model Parameter Settings}

The models, parameters, and their values we explored (model configuration) are:
\begin{itemize}
\item \textbf{Prompts}: Instruction Prompt (IP), RP-v1, RP-v2
\item \textbf{Text split}: max-token=4000, overlap-token: 40, char-per-token: 4
\item \textbf{Models}: qwen3:8b, gemma3:12b, gemma3:27b, llama3.1:8b, mistral:7b,  deepseek-r1:8b, meditron:7b, medllama2:7b
\item \textbf{Options}: context size window=32768, temperature=0.8, top-k=40, top-p=0.9
\end{itemize}




Among these parameters, higher temperature allows LLMs to generate more diverse or creative outputs, but also increase the chance of nonsensical, irrelevant, or factually incorrect outputs. 
We empirically verified that this did not significantly reduce extraction faithfulness due to human validation.
Although metaphor extraction is primarily an extractive task, a moderately high temperature was used to allow models to detect less explicit metaphorical expressions that may not follow fixed linguistic patterns. Human validation ensured that only faithful extractions were retained.

\subsection{Evaluation Setting}

In addition to the automatic verification steps shown in Figure~\ref{fig.framework}, we conducted a human qualitative evaluation of the extracted metaphors.

Three native Dutch speakers with PhD degrees and expertise in computational linguistics, health communication applications, or related fields independently reviewed the system outputs. Each annotator was asked to assess whether each candidate metaphor satisfied the following criteria:

\begin{itemize}
\item \textbf{Faithfulness}: whether the metaphor was explicitly present in the original text, rather than hallucinated or paraphrased by the model;
\item \textbf{Metaphoricity}: whether the expression represented a genuine cross-domain conceptual mapping rather than literal language, idioms, or conventionalised expressions;
\item \textbf{Contextual appropriateness}: whether the metaphor reflected the intended meaning in the original context.
\end{itemize}

Each extracted item was labelled as \textit{valid metaphor} or \textit{invalid metaphor}. Disagreements were resolved through discussion among the annotators to reach consensus.

This human-in-the-loop evaluation served two purposes:

\begin{enumerate}
\item to ensure the linguistic validity and contextual correctness of the extracted metaphors;
\item to construct a high-quality metaphor dataset (HealthQuote.NL) suitable for downstream analysis and clinical communication applications.
\end{enumerate}

Only metaphors validated through this process were included in the final corpus and reported in the results.

\subsection{Quantitative Evaluation}

To assess the effectiveness of the prompting strategies, we measured the precision of metaphor extraction under different prompt settings.
Precision was defined as:
\[
Precision =\frac{Metaphor_{validated}}{Metaphor_{total}}
\]
\noindent where $Metaphor_{validated}$ and $Metaphor_{total}$ are the number of metaphors validated after human verification and the total number of candidate metaphors generated, respectively.

Table~\ref{tab:precision_extraction} shows the comparison between Instruction Prompt and Refined Prompts.

\begin{table}[h]
\centering
\small 
\begin{tabular}{lccc}
\hline
Prompt & Generated & V. & P. \\
\hline
Instruction Prompt & 72 & 41 & 56.9\% \\
Refined Prompt v1 & 38 & 24 & 63.2\% \\
Refined Prompt v2 & 174 & 24 & 13.8\% \\
\hline
\end{tabular}
\caption{Comparison of metaphor extraction performance across prompting strategies (V: validated; P: precision)}
\label{tab:precision_extraction}
\end{table}

Using the initial instruction prompt, the model generated 72 candidate metaphors across 13 interview documents. After human verification, 41 were confirmed as valid metaphors, resulting in a precision of 56.9\%. The remaining candidates included hallucinated metaphors, paraphrased expressions, and idioms.

The refined prompt (RP-V1) achieved substantially higher precision, confirming that structured prompting, chain-of-thought reasoning, and explicit extraction constraints reduce hallucination and improve faithfulness.

The refined prompt v2 generated substantially more candidate metaphors (174) due to broader model testing, but only 24 were validated as true metaphors, resulting in a precision of 13.8\%. This indicates that while providing the full English Metaphor Menu increased sensitivity, it also introduced significant \textit{noise} and \textit{over-interpretation} for some LLMs. In contrast, refined prompt v1 achieved the best balance between precision and extraction quality.

\subsection{Output with initial instruction prompt}
We first list some challenges with example outputs using llama3.1:8b and
only initial Instruction Prompt (I.InP) without CoT on interview data: 

1) Some \textbf{abstractive or paraphrased} summarised metaphors, instead of extractive: e.g. a) from Doc-6:
"Een strijd voeren voor je lijf" (Fighting a battle for your body) from Full sentence: "Dat doe je ook deels voor je familie. Want die vinden, ja die willen graag dat je blijft en dan zegt het lijf vanzelf 'ja ik ga knokken'." (You do it also partly for your family. Because they want to stay and then your body says itself "yes I'll fight"); and another one b) from Doc-12: "Het is alsof je in de nacht ziet en je weet niet wat er aan de hand is" (It's like seeing things in the dark and not knowing what's going on) from Full sentence: "Maar het was alsof ik in de nacht stond en niets zag. Ik wist niet wat er aan de hand was." ("But it was like standing in the night and seeing nothing. I didn't know what was going on.")

2) \textbf{Fail to translate} simple sentences: e.g. from Doc-8:
De zonsondergang van zijn leven" (``no direct English translation'' from LLM, shall be ``the sunset of his life'') but it gave an explanation of the metaphor:
``using sunset as a metaphor for the end of one's life''.

3) \textbf{Suggesting metaphors output} plain text: e.g. from Doc-9:
For the original text in Dutch: "Het is heel belangrijk dat patiënten heel positief blijven en de familie ook - niet alleen familie, vrienden- meepraten en vooral niet bang zijn" (translation in English: "It's very important for patients to stay positive and for the family too - not just family, friends - to discuss things and especially not be afraid"), LLM suggests the following metaphor: ``Comparing the patient's journey to a ship navigating through stormy waters. The speaker emphasises the importance of staying positive and discussing things openly.'', which can be  useful in communication to support patients illustratively.

4) \textbf{Figurative language} rather than strict metaphors: e.g. in Doc-9:
the original Dutch text "Ik denk dat we allemaal een beetje te vroeg geboren zijn" 
(translation in English: "I think we're all a bit too early born"). It was interpreted as metaphor by LLM ``Comparing human beings to products that are not yet ready for use. The speaker humorously suggests that they and their family members are still learning how to navigate the healthcare system.''
However, this is more as a figurative language expression, instead of metaphor mapping A to B.

5) \textbf{Over-Interpreting metaphors}: e.g. from Doc-10,
a metaphor "een klein schip in de storm" (English Translation: "A small ship in a storm") is suggested from ``het wachten'' ("the waiting"), although the context did not say much about this metaphor. Instead, the researcher was asking the patient about the concern of process, as quoted below.
``O: Heel goed, dank u. Wat in het hele traject vond u eng of vervelend?
P: Het wachten.
O: Wat waren met name de momenten van het wachten?
'' (translation: ``O: Very good, thank you. What did you find scary or annoying about the whole process?
P: The waiting.
O: What were the specific moments of waiting?'')

Despite these issues, we were able to extract and collect 28 metaphors from the 13 interviews after human verification using the initial instruction prompt.
They reflect some of the source domains we defined later for the refined prompts, including journey, religion, control, machine, nature, and violence.
From a functional perspective, they can help communicate: feeling, hospital, process, dealing, life. These also reflect our function design in RPs (Section \ref{sec:method-design}).

We also explored I.inP with verification tool (\textbf{I.inP.VT}) control and the original English Metaphor Menu webpage link (I.inP.V.L).
Overall, 7 were re-identified by the model, however, there are also 3 variations and 13 new ones generated, together with the original 28 resulting in \textbf{41 in total from the initial instruction prompts}.
For examples of variations, e.g. 
from Doc-1, the full sentence "P: Ja, die morfine maakt dat je, die die heeft me ook wel angstig gemaakt." ("P: Yes, that morphine makes you, that made me anxious too."), I.inP generated ``Morfine brengt angstgevoelens naar boven" 	(Morphine brings anxiety feelings to the surface), while the I.inP.VT generated more indirect and figurative expression ``Er zit een morfine achter je" (There's morphine behind you). 
The metaphor is that the morphine is a physical presence that's affecting their behaviour.

In the outputs using the verification tool and the English Metaphor Menu website page link, we have new findings that we list here following the aforementioned list.  

6) The LLM extracted \textbf{idioms instead of metaphors}. For instance, from Doc-1 
"Een baby-hoeveelheid" (A baby-amount) is extracted to express the dietary change from the patients, 
from the full sentence: "Het is maar een baby-hoeveelheid. Nog niet eens de helft van hij vroeger at." ("It's only a baby's amount. Not even half of what he used to eat.)

7) The LLM \textbf{hallucinated metaphors}, e.g. from Doc-5, LLM generates a metaphor "De dokter is een dominee" (The doctor is a preacher), from the original sentence: "Die arts-assistent was echt 0 komma 0." (English translation: "That doctor's assistant was 0 point 0."). LLM explains ``The speaker compares the doctor to a preacher, implying they were ineffective and unhelpful.''
Firstly, it switched the role from ``doctor's assistant'' to ``the doctor''. Secondly, ``preacher'' was not used in the context.
The difference between Point 5 and Point 7 is that, Over-interpreting metaphors is more than suggestive metaphors but still try to interpret the situation; however, Hallucinated metaphors even mis-interpreted the roles in the context, e.g. ``doctor'' to the ``assistant''.

\subsection{Output with Refined Prompts}
We applied Refined Prompts (RPs) with CoTs (v1 and v2) and English metaphor menu as inserted knowledge (full text, instead of webpage link), in-context learning for v2.
We applied these two RPs on the interview data, and the collected metaphors are categorised into types (word, phrase, sentence), source domains, and functions.

With RPs and inserted knowledge, \textbf{24 metaphors} are collected after human verification from collective LLMs, Table \ref{tab:model_counts}. The output metaphors from RPs are more closely related to the original context and have less hallucination. 
Thus, in total, the initial instruction prompts and RPs have \textbf{collected 65} (41+24) metaphors from interview data.

However, it is still not possible to totally avoid it. 
For instance, from Doc-3,
LLMs (mistral:7b) attempt to create a metaphor out of the original phrase "Volle melk, volle yoghurt.." ("Full milk, full yoghurt...").
It describes "The patient compares their current state to a full milk or yogurt container, possibly emphasising the feeling of being overwhelmed.", with "reasoning": "This is metaphorical because the context suggests an emotional burden that goes beyond physical nourishment."
However, the original context has only talked about food.


\begin{table*}[t!]
\centering
\begin{tabular}{lccccccc}
\hline
Overall & Qwen & Mistral & MedLlama & DeepSeek & Llama3.1:8B & Gemma3:12B & Gemma3:27B \\
\hline
24 & 5 & 8 & 1 & 0 & 5 & 1 & 4 \\
\hline
\end{tabular}
\caption{Validated metaphor counts from collective LLMs using RPs on interview data.}
\label{tab:model_counts}
\end{table*}

\subsection{Application to blog data}
As a pilot study, applying our automatic metaphor extraction method with Gemma to the first 100 blog posts on \textit{Kanker.nl} produced a diverse set of figurative expressions related to the lived experience of cancer. The raw output contained both \textbf{novel metaphors} and \textbf{frequent idiomatic expressions} that are part of everyday Dutch. Examples of these \textbf{conventionalised or dead metaphors} include \textit{``te horen krijgen''} (``to be told''), \textit{``meten is weten''} (``to measure is to know''), \textit{``ik baal als een stekker''} (``I’m as annoyed as a plug''), and \textit{``laten we hopen dat alles met een sisser afloopt''} (``let’s hope it all blows over quietly''). While these idioms are emotionally expressive, they are generic and not specific to the cancer experience.

In total, out of the 100 posts, the model (\textit{gemma3:27b, olllama})  identified 39 posts that contained at least one metaphorical expression, yielding \textbf{65 distinct metaphor} instances. 
Of these, approximately \textbf{ten} were identified as particularly vivid and conceptually rich, reaching a level that could be considered for inclusion in a future \textit{metaphor menu}--a curated set of expressive metaphors for therapeutic or communication purposes. These most salient examples demonstrate how patients and relatives employ creative language to express emotion, reframe vulnerability, and negotiate meaning in relation to illness.

During the {manual post-filtering} phase, the analysis focused on \textbf{novel metaphors}: linguistically creative and context-specific expressions that reveal how patients conceptualise illness, identity, and recovery. The selected metaphors, grouped below by their \textbf{metaphorical vehicle}, illustrate how individuals on \textit{Kanker.nl} use figurative language to articulate emotion, reclaim agency, and construct meaning in the face of illness.\footnote{Note that we do not release patient's literal words here but rephrased them to protect patient privacy.}

\paragraph{The Party}
Here, cancer is personified as an uninvited celebrant. The party metaphor conveys irony and defiance, transforming fear into dark humor. This creative reversal allows the writer to express anger and regain narrative control by reframing the illness as something absurd rather than purely tragic.

\begin{quote}
\textit{``I was told not to use worrying titles anymore, so this time I’ve decided to make it a celebration instead -- my cancer is having a party of its own.”''} 
\end{quote}

\paragraph{The Car}
The damaged car serves as a vivid metaphor for physical and emotional wear. It conveys the experience of being battered but not beyond repair, blending vulnerability with resilience. The detailed imagery grounds the illness experience in tangible, everyday language that conveys fatigue and determination simultaneously.

\begin{quote}
\textit{``I feel like an old wreck of a car, the kind that can’t be patched up with filler anymore -- it needs entirely new bodywork, like the rusty 2CV sitting in someone’s garden.''} 
\end{quote}

\paragraph{The Lighthouse}
The lighthouse represents guidance, stability, and safety. It evokes the dependable presence of loved ones who provide light and orientation amid uncertainty. This metaphor articulates the relational and emotional support structures crucial for coping with illness.

\begin{quote}
\textit{``A lighthouse -- someone solid and constant, always there to guide you through the dark.''}
\end{quote}

\paragraph{The Storm}
The storm metaphor externalizes psychological turmoil and intrusive thoughts. By likening mental distress to turbulent weather, the speaker gives form to anxiety and confusion, making inner chaos visible and easier to communicate.

\begin{quote}
\textit{``It feels as if a wild storm is roaring through my head.''}
\end{quote}

\paragraph{The Train}
The train functions as a metaphor for relentless motion and loss of control. These images capture the unstoppable pace of medical treatment and the sense that life continues regardless of one’s readiness, balancing fatigue with perseverance.

\begin{quote}
\textit{``Just when you think you’ve stepped off one speeding train, you realize you’re already on the next one.”''}
\end{quote}

\begin{quote}
\textit{``Life keeps pounding along like a fast-moving train -- whether I’m ready to get on or not.''}
\end{quote}

\paragraph{The Painting}
Referring to the red surgical markings before an operation, the painting metaphor reinterprets the medicalized body as a canvas. It reframes vulnerability as something that can hold meaning and even beauty, allowing patients to reclaim agency over their altered bodies.

\begin{quote}
\textit{``My stomach looks like a work of art now, covered with careful red lines and circles drawn before surgery.''}
\end{quote}

\paragraph{The Final Chord}
The musical metaphor of the \textit{final chord} transforms death into a form of completion rather than cessation. It provides emotional and narrative closure, allowing grief to be expressed through harmony and resolution.

\begin{quote}
\textit{``This feels like the final note in the song of our journey together -- the moment of closure that sends everything gently into silence.''}
\end{quote}

\smallskip


These selected metaphors show how writers on \textit{Kanker.nl} employ creative figurative language to convey themes of control, damage, support, turbulence, transformation, and closure. The combination of automatic extraction and qualitative interpretation reveals how metaphor functions as a key tool for emotional expression and meaning-making in online narratives of illness.

\subsection{GPT on forum data}
\begin{figure*}[!t]
\begin{center}
\includegraphics[width=0.9\textwidth]{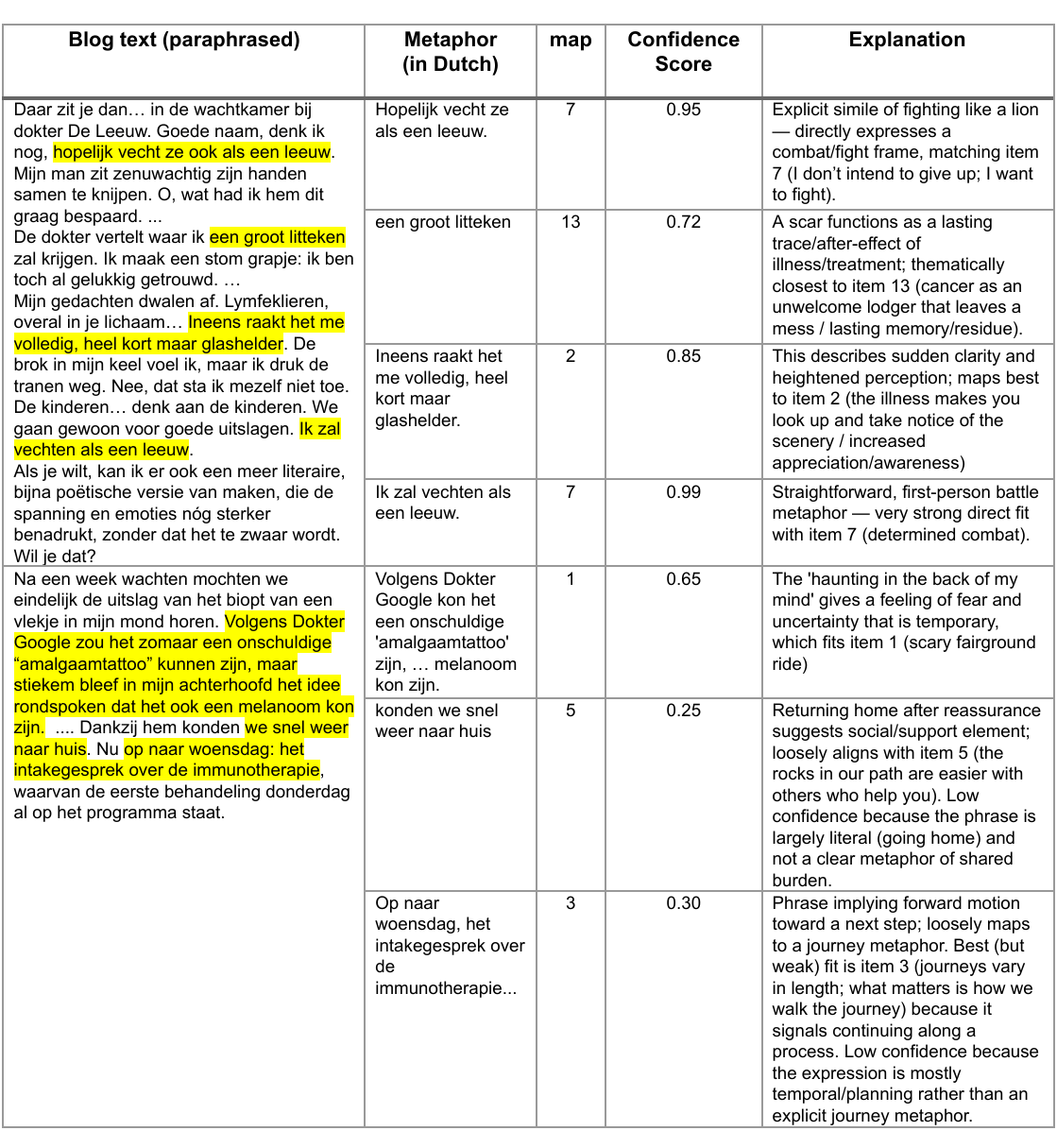}
\caption{Metaphor identification using GPT5 on forum data (paraphrased)  with model confidence scores: part-1 (`map' column refers to the value in original English Metaphor menu)
}
\label{fig.gpt-engligh-map-part1}
\end{center}
\end{figure*}

\begin{figure*}[!t]
\begin{center}
\includegraphics[width=0.9\textwidth]{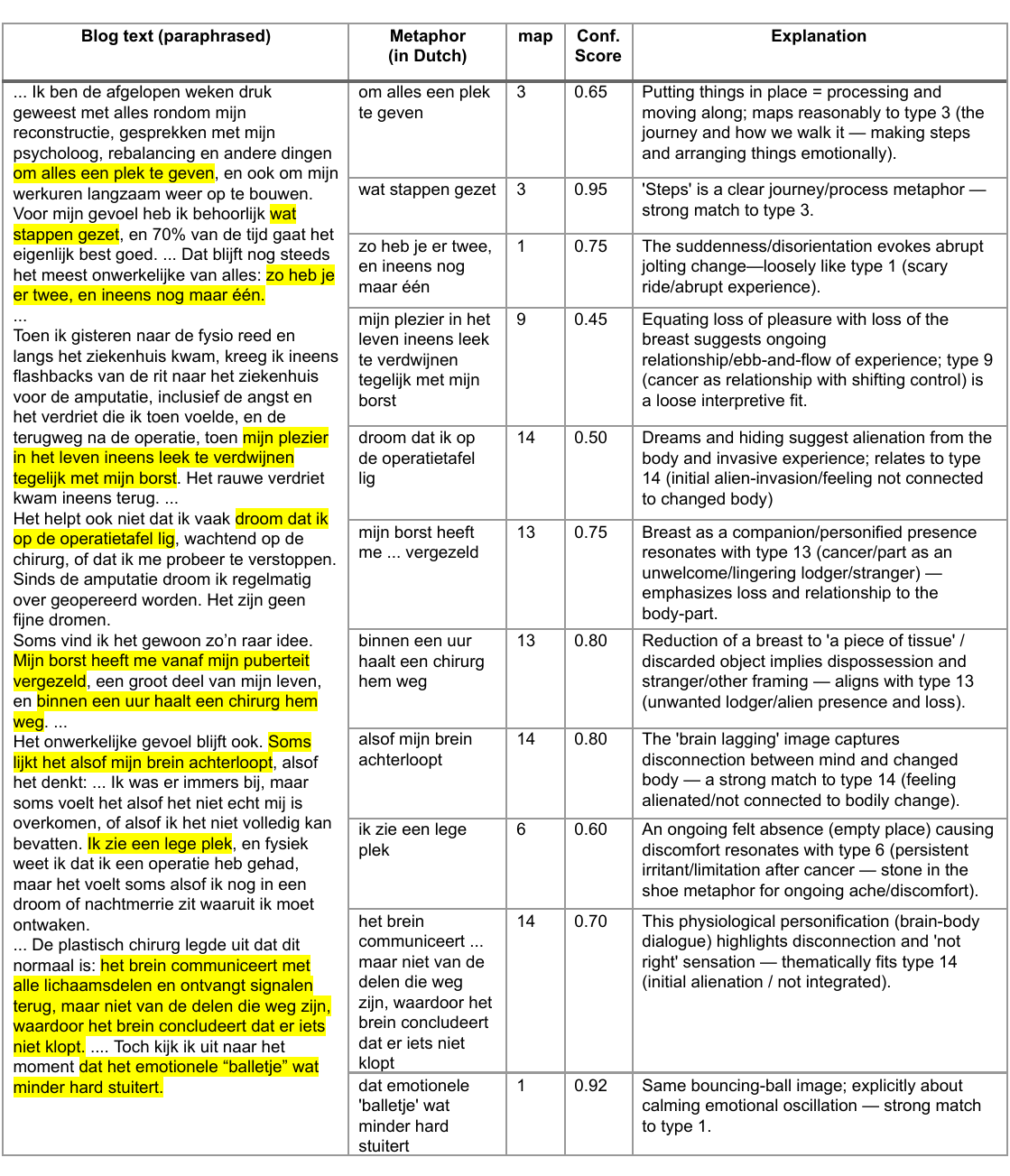}
\caption{Metaphor identification using GPT5 on forum data (paraphrased)  with model confidence scores: part-2 (`map' column refers to the value in original English Metaphor menu)
}
\label{fig.gpt-engligh-map-part2}
\end{center}
\end{figure*}

To compare with the open-source LLMs, we also used GPT models from OpenAI on the forum data for the metaphor extraction task and prompted GPT5 to map the extracted metaphors to the English Metaphor Menu and provide a confidence score. We list some of our extracted metaphors using GPT5 (\textit{paraphrased}) in this section, which align with the original English metaphor menu, as in Figures \ref{fig.gpt-engligh-map-part1} and \ref{fig.gpt-engligh-map-part2}.

\section{Discussion}

Firstly, without the CoTs and verification tools we implemented, the initial instruction prompts produced abstractive and suggestive metaphors, paraphrased from what was said in the interview data, instead of word-for-word extractions.
With the refined prompts and the inserted English metaphor menu knowledge, the LLM-generated metaphors are more closely reflecting the transcription content, but also less diverse or creative than the initial prompting outputs.

Regarding the corpus \textbf{HealthQuote.NL} we generated, it can be useful for multiple purposes, as a whole or a subset of columns.
For instance, users can focus on the Dutch metaphors column for Dutch-focused patient care support. 
The Dutch-English bilingual metaphors can be used for non-Dutch native speakers but who speak English in the care setting.
The categorisation, description, and reasoning of metaphors can be used to conduct a metaphor study itself and the relations within.
The prompts we used and the corresponding varied outputs from LLMs can be explored by AI and NLP practitioners who are interested in exploring prompt engineering for better analysis of non-literal language use.



For the phrase-level and word-level metaphors, we always add the context sentence as examples to the dataset for understanding and usage. 
For instance, at word level, the pair of original quote "balletje" and translation ``ball'', is accompanied with the context ``O: dan gaat het balletje heel snel rollen.'' (\textit{O: Then the ball starts rolling very quickly.}.)
We also give the description ``In this case “balletje” means ball, like a golf ball or football. Here it means that cancer treatment is like a fast rolling ball, so that’s the metaphor''.



% ...


From the interview transcriptions, there are many broken and incomplete sentences, e.g. from Doc-0, Original sentence: ``En, maar met [naam arts] nu, dat gaat een stuk beter en we hebben sinds dat wij, dat hij in [plaatsnaam] is gaan kuren, heb je zo’n oncologieverpleegkundige en daar kan je wel je vragen aan kwijt en zij... (\textit{But now with [name of doctor], it goes a lot better and since we've been at [place name] to cure, you have an oncology nurse there where you can put your questions and she...})''.
The metaphors we collected using two data sets are also different in the domain words that people used to map to (the vehicle).

From our experimental outputs, different LLMs can extract and suggest different metaphors from the same data we used even though the overlaps happen sometimes. So, it is useful to have a \textit{collective of LLMs} to carry out such a task from our findings. 

We have collected metaphors that patients can use not only in the cancer treatment process but also for hospitals, clinicians, and families, which can be very useful in a diverse way. 
From a clinical perspective, the extracted metaphors provide insight into how patients conceptualise illness, which can help clinicians tailor communication strategies. This aligns with prior work showing that metaphor-sensitive communication improves patient understanding and emotional support.

\section{Limitations}

This study has several limitations.
First, the dataset size remains relatively small, particularly for the interview data. While sufficient for exploratory analysis, larger datasets would enable more robust quantitative evaluation.
Although formal inter-annotator agreement metrics such as Cohen’s Kappa were not computed, all annotations were independently performed and disagreements were resolved through discussion to reach consensus.
Third, the prompting strategies were evaluated primarily on local LLMs. Results may differ for larger commercial models.
Fourth, distinguishing metaphors from idioms and conventionalised expressions remains challenging, both for LLMs and human annotators.

\section{Conclusions and Future Work}

In this work, we explored different kinds of prompting techniques for Dutch metaphor identification and extraction in cancer domain using two data sets including interview data transcription (oral language) and forum data (written language). 
Our findings provide practical guidance for prompt design in metaphor extraction tasks and demonstrate the feasibility of building metaphor datasets using local LLMs.
We collected 130 metaphors (65 each) from two data sets after human verification and mapped some of the metaphors to the original English Metaphor Menu.
Translation of metaphors between languages is challenging due to cultural and conceptual differences.
However, the collected metaphors cover all the categories that the original English metaphor menu has, with extended new types, and are further grouped into word, phrase, and sentence levels.
This collection of Dutch metaphors can be applied to support cancer patient care by integrating them into the care-path. 
For the current study, we only used 100 posts of the forum data we have. 
For future work, we will explore more data from the blog post for extended experiments.
We will look into some of the MWEs identified by LLMs, since they always overlap with idioms and metaphors \cite{hantowards}.
We also plan to explore the interpretability and explainability LLMs, incorporatin formal annotation protocols, and evaluating additional models. 

\section{Reproducibility}

To support reproducibility, we release:

\begin{itemize}

\item all prompts used in the study,
\item model configurations,
\item extraction framework code,
\item paraphrased and synthetic metaphor examples.

\end{itemize}

Due to privacy restrictions, original patient texts cannot be shared. The statistics of HealthQuote.nl is shown in Table \ref{tab:HealthQuote.nl-statistics}.

\begin{table}[t]
\centering
\small
\begin{tabular}{lc}
\hline
Dataset & Size \\
\hline
Interview documents & 13 \\
Blog posts analysed & 100 \\
Validated metaphors (interview) & 65 \\
Validated metaphors (blog) & 65 \\
Total validated metaphors & 130 \\
\hline
\end{tabular}
\caption{HealthQuote.NL dataset statistics}
\label{tab:HealthQuote.nl-statistics}
\end{table}

\section{Ethical Statement}
For both storytelling data (interviews) and kanker.nl data, we have gained ethical approvals within the research project to use. 
For the forum data, we used the secure version of OpenAI GPT5 established by the agreement between the University and OpenAI on anonymised and paraphrased data.
\section{Acknowledgement}
We thank Ida Korfage, Sheila Payne, Judith Spek, for the feedback on the abstract of this work.
We thank Richele and Iris Su Yi Tamminga from USC for their valuable discussion and feedback on the extracted metaphors.
Funded by the European Union under Horizon Europe Work Programme 101057332. Views and opinions expressed are however those of the author(s) only and do not necessarily reflect those of the European Union or the European Health and Digital Executive Agency (HaDEA). Neither the European Union nor the granting authority can be held responsible for them.
The UK team are funded under the Innovate UK Horizon Europe Guarantee Programme, UKRI Reference Number: 10041120.

\section{Bibliographical References}\label{sec:reference}

\bibliographystyle{lrec2026-natbib}
\bibliography{lrec2026-example}

\section{Language Resource References}
\label{lr:ref}
\bibliographystylelanguageresource{lrec2026-natbib}
\bibliographylanguageresource{languageresource}

\end{document}